\RequirePackage{fix-cm}
\documentclass[natbib,smallextended]{svjour3}       

\smartqed  

\usepackage{amsmath,amssymb,amsfonts}
\usepackage{algorithmic}
\usepackage{graphicx}
\usepackage{textcomp}
\usepackage{algorithm}
\usepackage{algorithmic}
  \usepackage{rotating}
        \usepackage{multirow}
        \usepackage{lscape}

\usepackage{color,soul}

\usepackage{xcolor}
\usepackage{caption}

\usepackage{subcaption}

\begin{document}

\title{Artificial Intelligence for Emotion-Semantic Trending and People Emotion Detection During COVID-19 Social Isolation 
}


\author{Hamed Jelodar  $^1$ $^2$  \and Rita Orji  $^2$   \and Stan Matwin  $^2$ $^4$   \and Swarna Weerasinghe $^3$   \and Oladapo Oyebode  $^2$   \and Yongli Wang  $^1$ }


\institute{ H. Jelodar \at
              jelodar@njust.edu.cn\\\\
              R. Orji \\
              rita.orji@da.ca\\\\
              S. Matwin \\
              stan@cs.dal.ca\\\\
              S. Weerasinghe \\
              swarna.weerasinghe@dal.ca\\\\
              O. Oyebode\\
              oladapo.oyebode@dal.ca\\\\
              Y. Wang \\
              yongliwang@njust.edu.cn\\
           \and
             \\
              $^1$School of Computer Science and Technology, Nanjing University of Science and Technology, Nanjing 210094, China \\\\
              $^2$Faculty of Computer Science, Dalhousie University, Halifax, NS, Canada \\\\
              $^3$ Faculty of Medicine, Dalhousie University, Halifax, NS, Canada  \\\\
              $^4$ Institute of Computer Science Polish Academy of Sciences, Warsaw, Poland   
                \at      
}

\date{Received: date / Accepted: date}

\maketitle

\begin{abstract}
Taking advantage of social media platforms, such as Twitter, this paper provides an effective framework for emotion detection among those who are quarantined. Early detection of emotional feelings and their trends help implement timely intervention strategies. Given the limitations of medical diagnosis of early emotional change signs during the quarantine period, artificial intelligence models provide effective mechanisms in uncovering early signs, symptoms and escalating trends. Novelty of the approach presented herein is a multitask methodological framework of text data processing, implemented as a pipeline for meaningful emotion detection and analysis, based on the Plutchik/Ekman approach to emotion detection and trend detection. We present an evaluation of the framework and a pilot system. Results of confirm the effectiveness of the proposed framework for topic trends and emotion detection of COVID-19 tweets. Our findings revealed Stay-At-Home restrictions result in people expressing on twitter both negative and positive emotional semantics (feelings), where negatives are “Anger” (8.5\% of tweets), followed by “Fear” (5.2\%),  “Anticipation” (53.6\%) and positive emotional semantics are “Joy” (14.7\%) and “Trust” (11.7\%). Semantic trends of safety issues related to  staying at home rapidly decreased within the 28 days and also negative feelings related to friends dying and quarantined life increased in some days. These findings have potential to impact public health policy decisions through monitoring trends of emotional feelings of those who are quarantined. The framework presented here has potential to assist in such monitoring by using as an online emotion detection tool kit.

 \keywords{Twitter, NLP, Deep Learning, COVID-19, Emontion}
\end{abstract}

\section{Introduction}
Over 73 million people have been affected by COVID-19 across the globe [3]. This more than a yearlong outbreak is likely to have a significant impact on mental health of many individuals who lost loved ones, who lost personal contacts with others due to strictly enforced public health guidelines of mandatory social segregation. Complex psychological reactions to COVID-19 regulatory mechanisms of mandatory quarantine and related emotional reactions has been recognized as hard to disentangle [1] – [4].  A study conducted in Belgium found social media being positively associated with constructive coping for adolescents with anxious feelings during the quarantine period of COVID-19 [4]. Another study conducted among social media users during COVID-19 pandemic in Spain was able to capture added stress placed on people’s emotional health during the pandemic period [5]. However,  social media providing a platform of risk communication and exchange of feelings and emotions to curb social isolation, this text data provides a wealth of information on the natural flow of people’s emotional feelings and expressions [6]. This rich source of data can be utilized to curb the data collection barriers during the pandemic. The goal of this research was to use AI to uncover the hidden, implicit signal related to emotional health of people subject to  mandatory quarantine,  embedded in a latent manner in their twitter messages.\\
 
Within the context of this paper, an NLP-based emotion detection system aims to provide useful information by examining unstructured text data  used in social media. The purpose of the NLP system used herein is to show the meaning and emotions of users' expressions related to a particular topic, which can be used to understand their psychological health and emotional wellbeing. In this regard, use of NLP-based approach for emotion detection from complex textual structures such as social media (e.g., Twitter) remains a challenge in biomedical applications of AI.\\

The goal of this paper is to contribute an AI based methodological framework that can uncover emotion semantic trends that can better understand the impact and the design of quarantine regulations. The two-fold objectives of this paper are: (a) to develop and AI framework based on machine learning models with for emotion detection and (b) to pilot this model on unstructured tweets that followed quarantine regulation using stay at home messaging during the first wave of COVID-19. We investigate emotions and semantic structures to discover knowledge from general public tweeter exchanges.  We analyze the structure of vocabulary patterns used on Twitter with a specific focus on the impact of the Stay-At-Home public health order during the first wave of the COVID-19. The AI framework is described and its implemented pipeline pilot herein can be used in the emotion detection of social media information exchange during the second wave of COVID-19 and beyond to investigate the impact on any future public health guideline.\\

We aim to demonstrate the effectiveness of deep learning models for detecting emotions from COVID-19 tweets.  In addition, our findings will provide directions public health decision making on emotion trend detection over a four weeks period in relation to “Stay at Home” order.  
The contributions of this paper can be summarized as follows:

\begin{itemize}

\item A cleaned and standardized tweet dataset of COVID-19 issues is built in this research, and a new database of emotion-annotated  COVID-19 tweets is presented, and this could be used for future comparisons and  implementations of detection-systems based on machine learning models.

\item We design a triple-task framework to investigate the emotions in eight standard positions (explained in section II B) via Plutchik’s model using the COVID-19 tweets in which all three different tasks are complementary to each other towards a common goal.

\item We discover semantic-word trends via various models such as latent Dirichlet allocation (LDA) and probabilistic latent semantic analysis (PLSA). We aim to have a semantic knowledge discovery based on topic trends and semantic structures during the first wave of the pandemic, which provides an effective mechanism for managing future waves.

\item A deep learning model based convolutional neural network (CNN)) is presented for emotion detection from the COVID-19 tweets. To the best of our knowledge, this is the first attempt that detects emotion automatically for people’s reaction to stay at home during the pandemic based on the online comments, especially for \#StayAtHome.

\end{itemize}

This paper is organized into sections: (a) a review of literature on existing  models on emotion detection for social media pertaining to health online communities (section 3) (b) introduce a multi-tasks framework to COVID-19 emotion detection (section 4), (c)describe data collection of twitter and research experiment (section 5), (d) discuss the effectiveness of the presented AI framework and future research directions (section 6) with final section on the conclusions on findings of emotion detection during stay at home(section 7). 

\section{NOVELTY OF THE PROPOSED FRAMEWORK}
Although machine learning based emotion detection approaches have been proposed within social media text analysis with the context of COVID-19, there are still many challenges remained to be addressed. In this regard, most of the existing studies related to COVID-19, on Twitter, and other social media platforms were performed on a general public opinion, no research have specifically investigated emotions related to quarantine “stay at home” order, public health policy of social segregation, that is widely used across the globe.  Novelty of the methods used in this paper consists of a multi-task framework that can be directly applied to COVID-19 related mood discovery, using eight types of emotional reaction and designing a deep learning model to uncover emotions based on the first wave of the pandemic public health restriction of mandatory social segregation. We argue that the framework can discover semantic trends of COVID-19 tweets during the first wave of the pandemic to predict new concerns that may be associated with furthering into the new waves of COVID-19 quarantine orders and other related public health regulations. Our novel approach presented herein can help future public health crisis management in the new waves of the Coronavirus  pandemic.

Moreover, public health decision makers need to understand the temporal patterns of emotional reactions on the population when these public health regulatory measures are continued. To fulfill this need, we investigate the semantic topic and emotion trends to better understand the people's reactions from the initial wave of the pandemic.

\section{RELATED WORK}
NLP and Machine Learning has been used within the context of identifying the type of emotions in twitter texts. In this section, we provide a review of literature on recent emotion detection studies with focus on; Emotion detection in online health communities, Emotion-based Lexical models, Deep learning and machine learning, and Directions for Public health decision making using social media during COVID-19 related text analytics.\\
Emotion detection analytics through information retrieval and NLP as a mechanism have been used to explore large text corpora of online health community communications in psychiatry, dentistry, cancer and health and fitness. For example, a communication tool was introduced for mental health care to understand counseling content based on emotion detection and natural language processing using chat assistants [7] - [12]. Similar to the proposed approach in our work, a research analyzed messages in online health communities (OHCs) to understand the most prominent emotions in health-related posts and proposed a computational model that can exploit the semantic information from the text data [9]. They presented a dataset from a cancer forum with the six common emotions based on the Ekman model and investigated the most prominent emotions in OHCs. We proposed to use broader types of emotions using Plutchik's model that contains eight emotions.\\

In our previous work [13], sentiment and latent-topics techniques application to COVID-19 comments in social media shed light on the usefulness of NLP methods in uncovering issues directly related to COVID-19 public health decision making. We expect to extend the methodology, in this study, within our goal of extracting meaningful knowledge of emotional expression words from people’s reactions during mandatory quarantine using the StayAtHome hashtag on Twitter. This knowledge is essential as it can help decision makers to take necessary actions to control the adverse emotional effects of various public health policies, especially during the emerging waves of the pandemic. Clearly, negative emotional effects such as anger and fear can lead to negative social reactions. To the best of the authors’ knowledge, little research have been done to understand the emotional expression during mandatory quarantine, partly due to difficulties in collecting such personal level data during the pandemic. The authors of a study in India analyzed real-time posts on Twitter during COVID-19, and they were able to identify the expression of mood of the nation through their analysis [14]. Also, they developed a platform for viewing the trends in emotional change across the country during a specific interval. As a result, their system allowed users to view the mood of India on specific events happening in the country during COVID-19. Development of such a platform for Canada may be a far reaching goal of this study. This study presents the first step towards development of such online tools to monitor the moods and emotions to inform public health decision makers.

\section{METHODS}
This paper’s methods provide step-by-step approach to text data processing, emotion detection and intensity scoring, emotion semantic trends calculation and finally evaluation of the deep learning algorithm using training and testing data.     

\subsection{Multi-Task Framework}
In this section, we present a multi-task framework based on Plutchik’s Emotion model [15] and deep learning techniques to address aforementioned research objectives of this paper. Our approach includes three main tasks. Plutchik’s is an operationalization of Ekman [16]. The first task is to create models to investigate emotional reaction to mandatory restrictions of  Stay-At-Home using tweets. The second task, is to show how to discover semantic and emotional trends to obtain patterns depicted in the first wave of the pandemic for the 30 days period from April 28th to June 1st, 2020. Finally, the third task, a machine learning deep neural network is built as an emotion detection system that can be used for social media exchange data analysis during the quarantine period. Our framework, including these three tasks, is presented in Fig. 1.\\

1)	\#StayAtHome-Related Tweets and Data-Dimension Reduction:\\
Our inclusion criteria include only tweets related to the COVID-19 and Stay-At-Home order. Application of this inclusion criteria is a critical step because the quality of the input data directly affects the output.Lexical text analysis,  Data-Dimension Reduction, and NLP Preprocessing of data are necessary to clean the data by removing the noisy and inconsistent tweets and also analyzing the relevant data to identify relevant and appropriate information related to the topic of interest. For this purpose, four NLP techniques are used: sentence splitting and word tokenization, removing stop-words, HTML cleaning to remove unnecessary contents, removing of stop-words and hashtags, and stemming to remove prefixes and suffixes hence returning to the root. The main purpose of the text data cleaning process is to eliminate all tweets unrelated to the subject of our stay at home public health order. Each tweet passed through a set of filters that are created based on the points described above. As stated in the objectives of this paper we want to detect the emotions expressed in the tweets that are in English language, non-English tweets are eliminated from the dataset and the classifier is then trained using English tweets only.

2)	Task I: Emotion-detection of \#StayAtHome tweets:\\
To achieve our research goal Task 1 is the most important process for automatic detection of emotions from \#StackAtHome tweets. It is the first step towards the initial determination of the type of COVID-19 emotions, which also has a direct influence on Tasks 2 and 3.  We take advantage of the NRC-Word-Emotion [17] lexicon based on Plutchik’s Wheel of Emotions to perform this task, which is beneficial for standard scoring of the word-emotion association. In this research, three sub processing steps are carried out on every annotated tweet: (a) identifying type of emotion using Plutchik-theory and hence reproducibility is warranted, (b) assigning the emotion score obtained from the National Research Council Canada (NRC)/ NRC Emotion Lexicon [17] and (c) identifying the emotion and the maximum association score based on the scores computed according to the following rule. In part (b) for assigning the score, we calculate the total emotion association score as the sum of scores of the terms depicting higher scores for higher intensity of the emotion of the tweets (See Table I, example). The scores denote the intensity of the emotion for the COVID-19 tweets. By default, every emotion in the tweet will receive a value of 1. This value will be increased or decreased by the intensifier words used in the tweets. However, if there are multiple mentions of emotion then the intensity will have a higher score, as shown in Table I. In part (c) the maximum association score of a tweet represents the maximum score noted for any of the eight emotions. A tweet that does not associate with any emotion receives a score of zero (0), as showed in Table 1. This AI based emotion detection task uncovers emotion semantics with emotion valuation (strength) attached to each emotion lexicon in each of the tweets. \\\\\\\\

\begin{figure}
\centering
  \includegraphics[height=10.14cm,width=12cm]{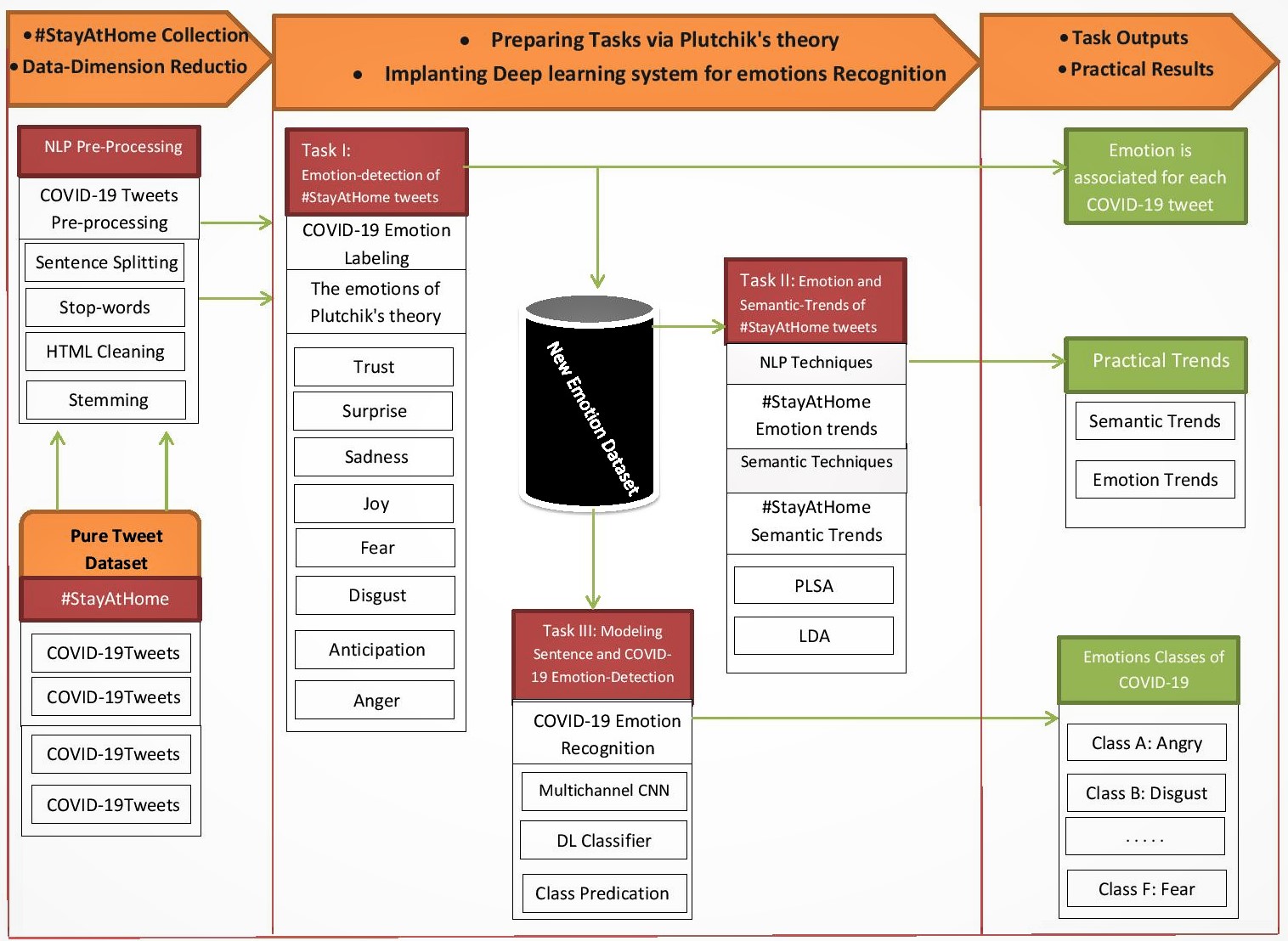}
\caption{Research framework and pipeline for the COVID-19 tweet emotion capture and analysis}
\label{fig:1}       
\end{figure}

\begin{figure}
\centering
  \includegraphics[height=7.14cm,width=12cm]{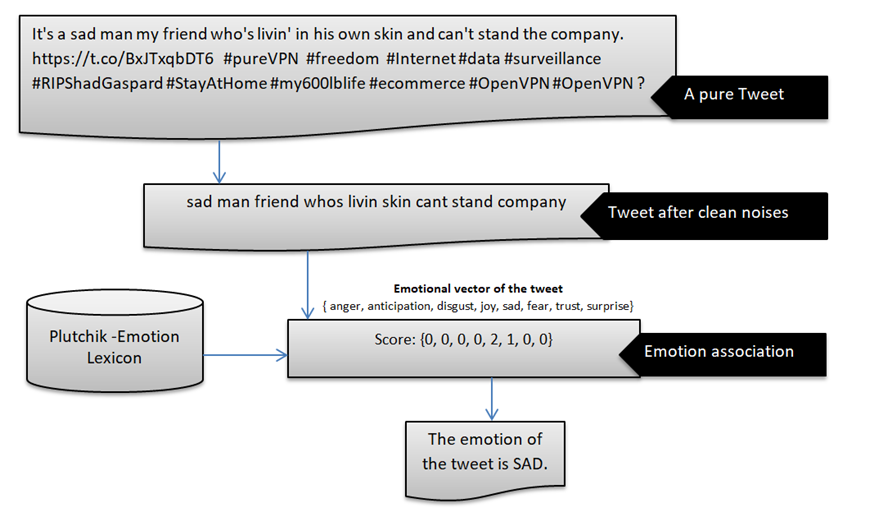}
\caption{Example of the process of determining the score for a pure tweet that is related to \#StayAtHome}
\label{fig:1}       
\end{figure}

Since the labelling is done automatically and no human tagging is used, we will have consistent data annotation. This model defines eight basic emotions and makes it possible to provide consistent classification of texts to uncover the trend in the data to reach the objectives of this research. Fig. 2 provides an example of selecting the score for COVID-19 tweets. For example, from the tweet after text processing  showed “Sad man friend whos livin skin cant stand company” and will have the emotion SAD associated with FEAR and this emotional expression provided the highest score from NRC based Lexicon when the process is detecting the predominant emotion. \\

\begin{table}[]
\caption{EXAMPLE OF THE TWEETS WITH VARIOUS EMOTIONS}

\resizebox{12cm}{!} {
\begin{tabular}{|l|l|l|l|}

\hline
\multicolumn{2}{|l|}{Tweet ID.} & Tweet without stop-words                                                                                                            & Label                         \\ \hline
\multirow{2}{*}{A}    & Tweet   & Today has been a challenging day, here's to tomorrow                                                                                & \multirow{2}{*}{Anticipation} \\ \cline{2-3}
                      & Score   & Anger=0$\sim$Anticipation=1$\sim$Disgust=0$\sim$Fear=0 $\sim$Joy=0 $\sim$Sadness=0$\sim$Surprise=0$\sim$Trust=0                     &                               \\ \hline
\multirow{2}{*}{B}    & Tweet   & A day is a long time in the coronavirus pandemic.                                                                                   & \multirow{2}{*}{Anticipation} \\ \cline{2-3}
                      & Score   & Anger=0$\sim$Anticipation=2$\sim$Disgust=0$\sim$Fear=0 $\sim$Joy=0 $\sim$Sadness=0$\sim$Surprise=0$\sim$Trust=0                     &                               \\ \hline
\multirow{2}{*}{C}    & Tweet   & Looking forward to those summer days when I can enjoy the beach and the ocean breeze again????. Stay positive and healthy everyone. & \multirow{2}{*}{Joy}          \\ \cline{2-3}
                      & Score   & Anger=0$\sim$Anticipation=1$\sim$Disgust=0$\sim$Fear=0$\sim$Joy=3$\sim$Sadness=0$\sim$Surprise=0$\sim$Trust=1                       &                               \\ \hline
\end{tabular}
}

\end{table}

3)	Task II: Emotion/Semantic-Trends of \#StayAtHome\\ 
Researchers have identified timing of the emotional progression and noted that positive emotions arose significantly earlier and the negative emotions took longer [18] – [20]. Identification of the emotion and semantic trends over time can be helpful and effective to understand temporal changes of the opinions related to the human behavior. In fact, understanding the mood changes or awareness of the emotion trends can have a practical application for public health decision making. We use semantic topics [21] discovered in the entire dataset to detect and describe semantic trends. \\

In order to obtain semantic topics we need to design a topic model. Infact, we consider two popular methods for evaluating and determining an optimal approach to obtain semantic-trends from \#StayAtHome tweets of the online community during the stay-at-home. For applying this task, the PLSA [21] and LDA [22] models are employed to obtain the best semantic related-words and discovering semantic structures of the COVID-19 tweets, as described below:\\
- The probabilistic latent semantic analysis (PLSA) model is used as an NLP technique that can display topical similarities between words.\\ 
- The latent Dirichlet allocation (LDA) model has proven very useful for semantic extraction and generating trends over time. LDA has been successfully applied in several applications such as topic discovery, temporal semantic trends, document classification, and finding relations between documents. Another advantage of this method is the identification of semantic-trends over time, which we consider in this research as means to discover unusual semantic trends based on the first wave of the pandemic of \#StayAtHome tweets. Overall, The main aim of the task 2 is to capture two kinds of trends based on emotion and semantic aspects of the COVID-19 tweets.   However, we know that LDA model discover ‘semantic related-words’ from the semantic structure of the text.  Then by investigating the distributions of these semantic-topics in various days, we obtain semantic trends.\\

As a second part of this task, we compute the types of each emotion to identify the trends among different emotions based on task I. By considering the length and strength of the staying at home public health order in the first wave, we believe that it is necessary to examine the changes in people’s emotions by monitoring the time trends and fluctuations of directions using twitter data. \\

4)	Task III: Modeling Sentence and COVID-19 Emotion-Detection\\ 
Machine learning offers the advantage of automatic emotion detection beyond using existing lexicial dictionaries for emotion analysis. In particular, deep learning models have proven successful in many NLP applications for emotion detection from health and medical text data [23] – [25]. We focused use a convolutional neural network (CNN)   [26] – [27] to implement an emotion detection system based on emotion vectors of \#StayAtHome tweets. Our network layers involve embedding layers, convolution layers, drop out layers, and Max-Pooling and filter layers.\\

In the following, we discuss the details of the designed deep-learning model, the number of convolutional layers, and dense layers to build our COVID-19 emotion detection framework; since the input layer is the sentence representation, a convolutional layer is then deployed to obtain the sequence level feature from the sentence sequences. Moreover, the convolutional layer is considered as the core functional block and includes a collection of filters. These filters serve as a learner in the network act when they find some certain type of feature at a determined input. Overall, we consider three flatten layers for the designed method. Finally, we concatenate the output of all three learned features by considering dense layers to generate scores and recognize the type of emotion of the COVID-19 tweets.

\section{EXPERIMENTAL  EVALUATION}
This section describes the dataset, generates emotion/semantic trends and the informative results with various experiments to evaluate the performance of the research model. In fact, to show the value of the framework, we conduct experiments on collected COVID-19 tweets and generate informative emotion/semantic trens.   We provide a comparison with a standard base-line to demonstrate the superiority of our CNN model for emotion detection using F1 score and accuracy as standard metrics.  In this research, we use 90\% of the data for training and 10\% for testing for all experiments. We focused on the tweets from the first waves of the COVID-19 pandemic based on \#StayAtHome. However, NLP pre-processing methods (such as stop-words and stemming) are used to reduce the noises and improve the quality of the output of the task. Table II shows various times along with the number of tweets based on pure content and status of data after removing duplicate contents.\\

\begin{table}[]
\caption{TABLE II. DETAILS OF THE \# STAYATHOME DATASET}

\resizebox{12cm}{!} {
\begin{tabular}{|l|l|l|l|}
\hline
\#StatAtHome Days & Pure Tweets & Obviate Duplicate Tweets & After Data-Reduction \\ \hline
28-4 To 6-5       & 409,000     & 145,000                  & 92,000               \\ \hline
7-5 To 15-5       & 289,000     & 112,000                  & 12,000               \\ \hline
16-5 To 24-5      & 226,000     & 88,000                   & 10,000               \\ \hline
25-5 To 6-2       & 122,000     & 54,000                   & 6,000                \\ \hline
Total             & 1,046,000   & -                        & -                    \\ \hline
\end{tabular}
}

\end{table}

\subsection{Informative Trends of the first wave: Emotion and Semantic}
Trending topics, to a certain extent, describe the opinion of a community and provide the means to analyze it, knowing where public attention is at a certain time point and this has become a matter of interest for researchers and health professionals. Regarding task II, we need to predict the trend of topics and give some explanations for the important variation of trends about COVID-19 trends. To test our machine learning approach with respect to this task, we randomly split our dataset into 90\% for training and the remaining 10\% for testing.

\subsection{Relationship between semantic trends and StayAtHome Tweets}
It is difficult to identify the key concepts discussed by users from a million tweets in traditional ways, so we examine NLP methods (LDA and PLSA) to extract topics based on semantic aspects to better understand behaviors and People's reactions, during stay at home. Then, we investigate the distribution of generated topics on different days of the initial wave of the outbreak, which as a result of this process can be helpful in managing public health in the community. First, we investigate PLSA and LDA models to analyze and validate the relationship between semantic-topics extracted from COVID-19 tweets and related issues of the pandemic. For this purpose, we use a Mallet package. Then, we generate 100 topics and only focused on 5 top topics of all COVID-19 tweets as a result of topic modeling for discovering semantic related-words. Fig 3, compares the performance of the two topic modeling methods we have considered to identify emotional trends in our twitter data.  And showed, LDA can capture semantic topics better than the PLSA model for extracting semantic-topics of \#StayAtHome tweets. Therefore, we consider an LDA model for performing Task II.

\begin{figure}
\centering
  \includegraphics[height=10.14cm,width=12cm]{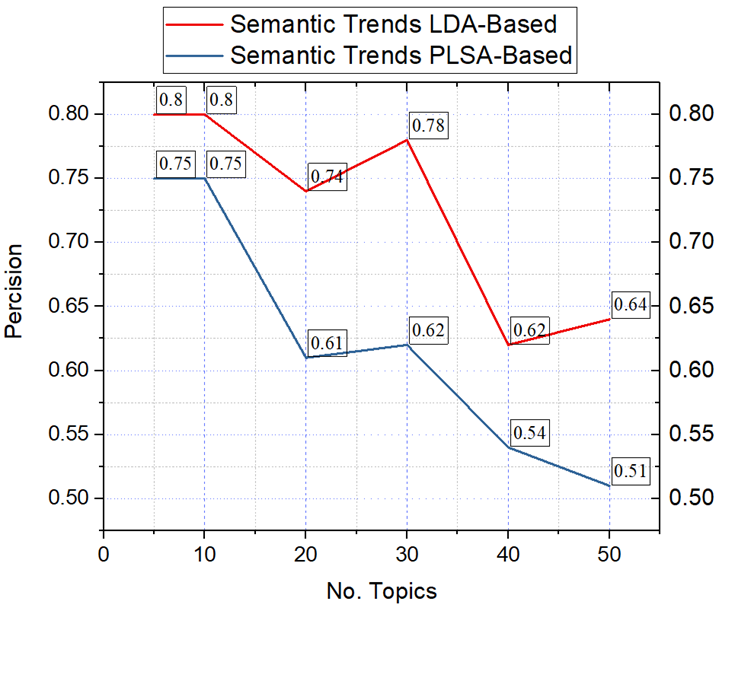}
\caption{The precision metric of word clustering for semantic-topic trending based on LDA and PLSA}
\label{fig:1}       
\end{figure}

To implement our analytic framework’s detection of semantic trends shown in the topics during the initial wave of the pandemic, we investigate five top topics (i.e., S1, S2, S3, S4, and S5, Fig.4) to better understand the online community reactions change over time. These topics are distributed over different days and we were able to isolate time varying nature of the semantic trends of \#StayAtHome tweets labeled by an automatic process described in Task 1. According to Fig. 4, the highest ranked (most frequent) topic is characterized by the words Home, Staysafe, Lockdown, Love, and Family.  These correspond to the safety issues related to staying at home. We label this topic as S1. It rapidly decreases over time at the rate of  0.11 (p=0.04) within the 28 days and the decline was greater within the last 14 days 0.28 (p=0.001) (Fig. 4) with some day to day fluctuations shown in the graph. Topic S2( words Live, Free) shows a decline within the first 14 days and then an increasing trend is detected within the last 14 days. We need to confirm these with data related to quarantine and it is possible that after the 14 day quarantine period the individuals feel free to live. It is important to notice that negative feelings grow over time, eg S4 (words Friends, Die, Virus) increases at a rate of 0.14 (p=0.0001) and  S5 (Home, remote, quarantine, health life) increases at a rate of 0.06 (p=0.0005) over the course of the 28 days of tweet text data collection period. Herein we show dynamic behaviour of statistically significant trends of topics from April 27th to June 1st.

\begin{figure}
\centering
  \includegraphics[height=10.14cm,width=12cm]{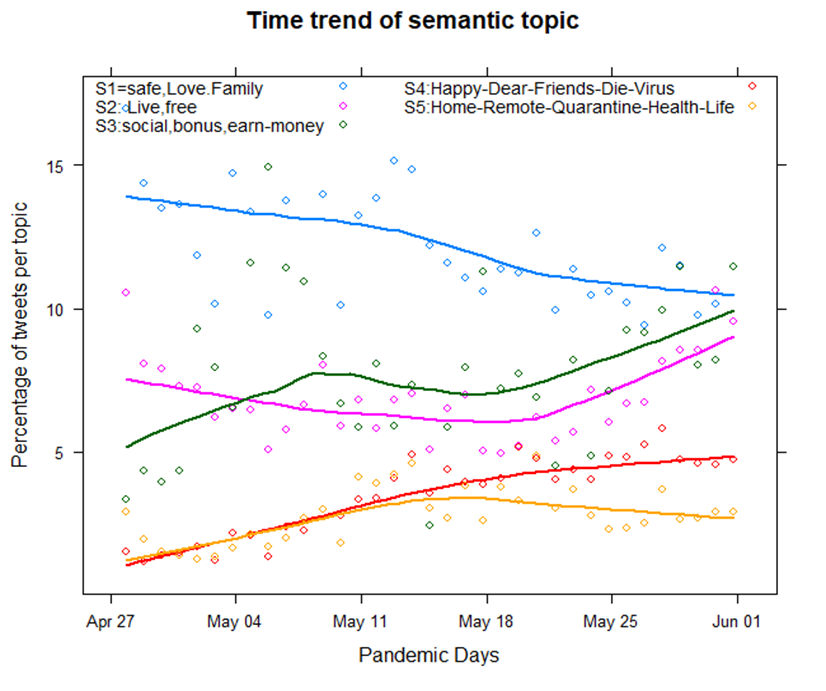}
\caption{Semantic trends of the initial waves of COVID-19 pandemic by \#StayAtHome}
\label{fig:1}       
\end{figure}

\subsection{Relationship Between Emotion Trends and StayAtHome Tweets}
In Tasks 1 and 2 of the framework, we take the advantage of NRC emotional lexical, which is supported by Plutchik’s theory based on 14,000 words for finding the eight primary emotions: anger, anticipation, joy, surprise, sadness, disgust, trust, and fear. The NRC dictionary has been widely validated for emotion analysis on social media such as Twitter emotion; therefore, we consider the advantage of this lexicon for \#StayAtHome tweets in this research.  In the application of task II analysis in this stage, we have accomplished the identification of the relationship between emotion trends in COVID-19 tweets based on Plutchik theory of classification.

\begin{figure}
\centering
  \includegraphics[height=10.14cm,width=12cm]{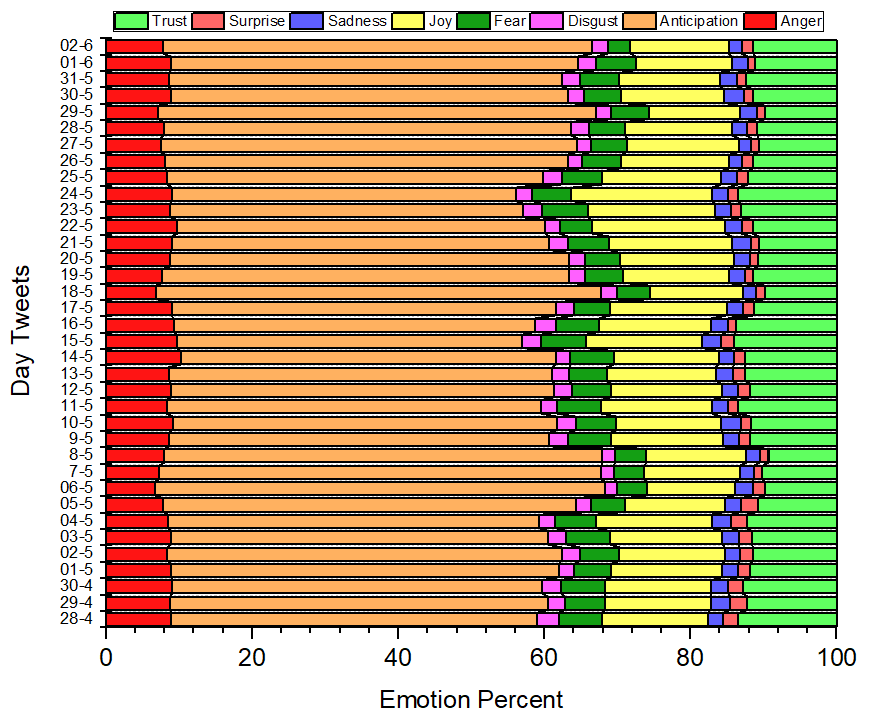}
\caption{Distribution of emotion trends in \#StayAtHome tweets over time in the initial wave of COVID-19  pandemic.}
\label{fig:1}       
\end{figure}

\begin{figure}
\centering
  \includegraphics[height=10.14cm,width=12cm]{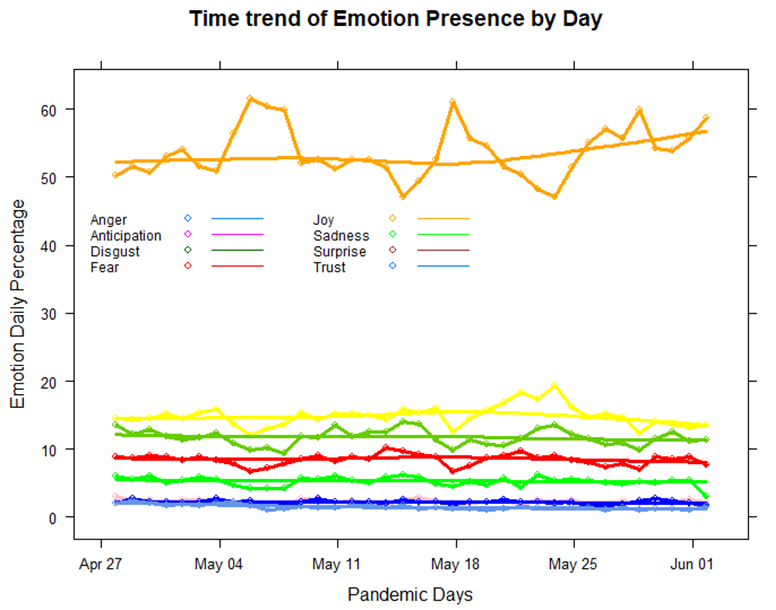}
\caption{Emotion Trends of the COVID-19 tweets in line graphs.}
\label{fig:1}       
\end{figure}

According to Fig. 5 the most of the emotions depict in tweets across time are “Anticipation”. As shown in Fig. 6 the mean percentage of “anticipation” detected per day across 30 day period is 53.6 (95\% CI: 52.4-55.0) with the least shown by “Surprise” (mean 1.5\% , CI: 1.4-1.6\%), and “Sadness” (mean 2.2,CI:2.1-2.3\%). Among negative emotions “Anger” is shown  with the highest (mean=8.5,CI:8.2-8.7\%), followed by “Fear” (mean=5.2, CI:5.0-5.5\%), “disgust” (mean=2.3, CI:2.2-2.4) and “Sadness” (mean=2.2, CI:2.1-2.3). Among positive feelings “joy” (mean=14.7,CI:14.4-15.4) is the second highest emotion.\\\\\\\\\\

According to psychology literature [36],  Anticipation and Surprise can be related to positive or negative health emotional outcomes. Nevertheless, in this study anticipation stemmed out of the hashtag “stay-at-home”, a restriction on a socially undesirable action and therefore, one can assume anticipation is mostly directed towards a negative emotional feeling, of perceived susceptibility. It is important to consider that these tweets were exchanged during the early pandemic period of April 28th to June 1st, 2020 of the first wave and North America reported the peak in May, 2020. Anger may be directed towards missing summer outdoor activities due to stay-at-home restrictions, whereas fear may be expressed by those living in high risk clusters of elderly and those with chronic conditions.   
It is important to note that the negative feeling of disgust was minimal and people may be aware of the importance of quarantine regulations. These emotion expressions and trend detection over time provide important messages where public health decision makers can be aware of in the future public health regulation ordering. Though people trust the public health measures, their anticipation towards negative emotions need to be considered in the way the public health regulations are ordered and imposed. Negative connotation of anticipation and fear can be overcome with public education using social media.

\subsection{Deep learning model configurations and Training details}
 The objective of task III of this work is to automatically detect emotions from \#StayAtHome tweets by enabling  Multi-Channel CNN methodology as a computational model for the emotion detection of the COVID-19 tweets.  The mode trained on tweets based on the created dataset of COVID-19 emotion (task II).  First we get the input data of the Task I, we leverage our COVID-19 tweet data to train our own word embedding with Word2Vec technique [13] which provides a much richer text representation than the classical, word-based approaches. The corpus for training word2vec is generated by selecting all the necessary words, followed by preprocessing the data and removing all stop words. The output of word2vec are real values are vectors (we used 100 as the dimension). For our implementation, we used to Keras library. Then  the dense layers are used to output the result of the deep learning model for  COVID-19 emotion detection. As already discussed, every training tweet is automatically annotated (labelked) using one of eight different emotions including fear, joy, trust, sadness, anger, surprise, and anticipation. To the best of our knowledge, this is one of the few studies of automatic emotion detection from COVID-19 tweets by focusing on Stay-At-Home issues.

\subsection{Validation and comparison of deep learning model}
As Long short-term memory (LSTM) [28] – [29] is a standard base-line for this research area, we also use that deep model in our study. We compare the CNN model discussed in sec. IV  with the LSTM-Softmax for COVID-19 emotion detection of the \#StayAtHome dataset. The LSTM network used in this research consists of 64 units. Here, we consider various parameters to train our model with the different number of epochs such as 10, 20, 30, 40 and 50 to ensure the significance of the obtained results. However, for each COVID-19 tweet, we have 8 labels that are features for our detection. Therefore, the output of the deep-learning model can determine the type of COVID-19 tweets with the labels.

\begin{figure}
\centering
  \includegraphics[height=8.14cm,width=12cm]{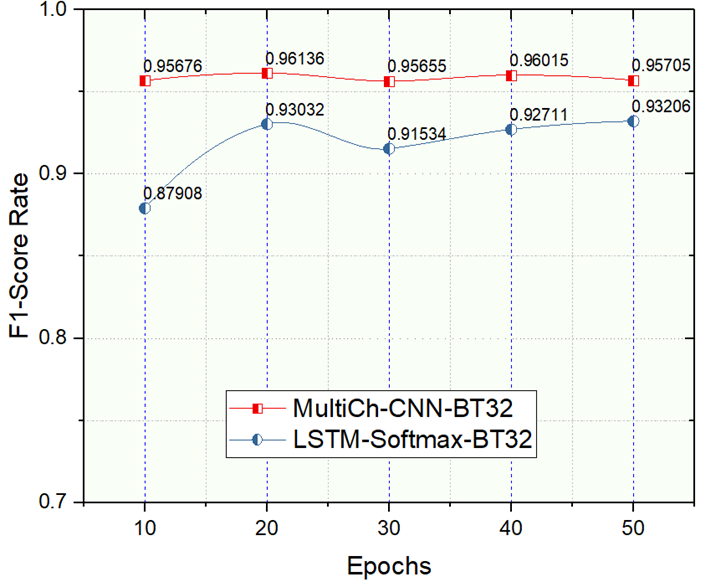}
\caption{The F1-score for COVID-19 emotion detection by comparing the CNN model and the LSTM model}
\label{fig:1}       
\end{figure}

\subsection{Evaluation of the effectiveness of the framework }
We evaluated the performance of the research model with emotions classes. Fig. 7 provides a clear view of variation with different parameters using word embedding trained. In particular, our preliminary results indicate that the multi-channel CNN out performs the  LSTM-Softmax in terms of various epochs based on a multi-class F-score as a standard metric. The advantage of the CNN model comes  for detection of the type of emotions in COVID-19 tweets, which enables to avoid overfitting and still be able to find complex patterns to emotion detection in the introduced data.

\section{DISCUSSION, LIMITATION AND FUTURE WORK}
Our results, in general, suggested that the machine learning methods we use are appropriate for the emotion detection of COVID-19 tweets. The study results clearly demonstrated anticipation as a prominent emotional semantic. Among various definitions for this emotion semantic analysis, anticipation is considered as one of “the mature ways of dealing with real stress”[30].  Regarding this definition, people can lower their stress during the COVID-19 pandemic by anticipating and preparing how they are going to deal with it. Anticipation can be interpreted as either future positive and negative events, according to [31], and are aligned with hope and fear which are the typical anticipatory feelings that arise in response to possibilities of future such events. A study that included multiple unigrams and bi-grams related to COVID-19 twitter feeds were analyzed using machine learning approaches and their findings were similar to ours in that the dominant theme identified was anticipation with a mixed of feelings of trust, anger and fear [32].\\ 

To develop a framework that can understand the type of standard emotions contained in COVID-19 sentences in social media is among the challenging topics of NLP for the public health and mental healthcare delivery [33] – [35]. Therefore, in this paper, a multi-task framework is presented to make a smart emotion detection system based on the Stay-At-Home aspects of COVID-19 tweets. All the experiments are performed using different parameter settings. The results suggest that the CNN model with two convolutional layers with filter sizes of 3 and 4 can achieve good performance with various metrics for emotion detection and classification.\\

Use of online social network text data to understand user health behaviors and emotions has become an emerging research area in NLP and health informatics [36] – [49]. COVID-19 introduced an unprecedented global threat that public health planning and policy making community are still struggling to find best practices to curb the pandemic. As the pandemic evolved public health guidelines became strict measures imposed on the general public. This one track minded approach of combating the spread or better known as flattening the curve, neglected emotional and mental health of the individuals who were subject to those strict public health ordering. This study findings showed a mechanism of how the emotions and semantic trends of people’s reactions to COVID-19 public health restrictions can be obtained for knowledge discovery and can inform related decision making. The advantage of such an approach is that identifying these online trends provide easy and helpful information about public reactions to  particular issues and thus it has recently attracted the attention of medical and computer researchers. The framework proposed in this research covers three practical tasks that are related to each other with a common goal to develop a deep-learning system for emotion detection and analysis of informative trends from COVID-19 tweets of people’s reaction during the stay-at-home. Our final results uncovered important directions for public health policy makers [40] and decision makers [41] to pay attention to emotional issues that stemmed from those strict public health restrictions. \\

This research has some limitations, e.g the size of dataset, data inclusion limited to emotions based on texts of COVID-19 issues. Currently, our data consists of 1,047,968 tweets based on \#StatAtHome tweets from 28-April to 2-June of 2020. Although more tweets can be extracted based on \#StayAtHome, we believe that the number of current tweets is sufficient to draw reasonable conclusions to direct possibilities of uncovering importance of consequences of public health orders and restrictions. We acknowledge that the imbalanced dataset representing different emotions is also a limitation of our work. We did not consider slang  or emoticons to compute emotions in the tweet contents, it would be useful to build a new emotional lexical to cover slang words related to COVID-19 issues. A significantly longer temporal horizon longitudinal dataset would allow us  using LSTM on sequences of tweets, as well as replacing the cross-validation. train/test approach with one based time-stamp rather than the cross-validation approach. Further advantage of a temporally larger dataset is an opportunity of a longitudinal study combining geographically-based tweeter -detected emotions with COVID-19 incidences and expanded public health regulations to enable geographic-area targeted public health decision making. The framework we developed showed potential to accurately uncover emotional responses and temporal trend detection of mood changes due to quarantine related public health orders.

\section{CONCLUSION}
This paper presented a novel framework  for emotion detection using COVID-19 tweets in relation to the “stay-at-home” public health guidelines. For this framework, a multi-task framework of COVID-19 emotions detection via a CNN model was presented. The research further shows that the framework is effective in capturing the emotions and semantics trends in social media messages during the pandemic. Moreover, it presents a more insightful understanding of COVID-19 tweets by automatically identifying the type of emotions including both negative and positive reaction and the magnitude of their presentation. The framework can be applied to uncover reactions to similar public health policies that affect people’s well being. We identified ways to improve the findings in future research. We discuss potentially significant, realistic future work, such as   extending the longitudinal character of the results, inclusion of geography-based public health orders and spatially-annotated COVID-19 case loads.

\section*{Ethical Approval }
All procedures performed in studies involving human participants were in accordance with the ethical standards of the institutional and/or national research committee and with the 1964 Helsinki declaration and its later amendments or comparable ethical standards.\\\\\textbf{Declaration of Conflict of Interest :} All authors declare no conflict of interest directly related to
the submitted work.\\

 \nocite{*}


\end{document}